\documentclass{article}
\usepackage{ijcai21}

\usepackage{times}
\usepackage{soul}
\usepackage{url}
\usepackage[hidelinks]{hyperref}
\usepackage[small]{caption}
\usepackage{graphicx}
\usepackage{amsmath}
\usepackage{subfigure}
\usepackage{multirow}
\usepackage{amsthm}
\usepackage{booktabs}
\usepackage{algorithm}
\usepackage{algorithmic}
\usepackage{amssymb}
\usepackage{color}

\title{Video Salient Object Detection via Adaptive Local-Global Refinement}

\author{
Yi Tang$^1$
\and
Yuanman Li$^{2}$\and
Guoliang Xing$^1$
\affiliations
$^1$The Chinese University of Hong Kong\\
$^2$Shenzhen University\\
\emails
\{yitang, glxing\}@cuhk.edu.hk,
yuanmanli@szu.edu.cn
}

\begin{document}

\maketitle

\begin{abstract}
	Video salient object detection (VSOD) is an important task in many vision applications. Reliable VSOD requires to simultaneously exploit the information from both the spatial domain and the temporal domain. Most of the existing algorithms merely utilize simple fusion strategies, such as addition and concatenation, to merge the information from different domains. Despite their simplicity, such fusion strategies may introduce feature redundancy, and also fail to fully exploit the relationship between multi-level features extracted from both spatial and temporal domains. In this paper, we suggest an adaptive local-global refinement framework for VSOD. Different from previous approaches, we propose a local refinement architecture and a global one to refine the simply fused features with different scopes, which can fully explore the local dependence and the global dependence of multi-level features. In addition, to emphasize the effective information and suppress the useless one,  an adaptive weighting mechanism is designed based on graph convolutional neural network (GCN). We show that our weighting methodology can further exploit the feature correlations, thus driving the network to learn more discriminative feature representation. Extensive experimental results on public video datasets demonstrate the superiority of our method over the existing ones.  
	.
\end{abstract}
\section{Introduction}
The purpose of salient object detection (SOD) is to focus on the most attractive objects or regions in an image or video and then highlight them with unambiguous boundaries from the complex background \cite{wei2020f3net}. Due to the trait of this community, it is usually treated as the pre-processing to support other visual tasks, such as visual tracking, image retrieval and so on. 

As for the approaches of salient object detection, we can roughly divide them into two categories by the input of algorithms, i.e., image salient object detection (ISOD) and video salient object detection (VSOD). In this paper, we only focus on the latter one.
Compared with the ISOD, the saliency detection in video sequences is to capture the motion cues by the sequential inputs, because moving objects are more likely to attract human's attention and become the salient objects. However, saliency shift \cite{Fan_2019_CVPR} is widespread in video sequences. It means that the salient objects may gradually change in a long video. In order to solve this issue, existing methods need to capture both static cues and motion cues. In other words, the approach is to synchronously extract the intra-frame spatial features and inter-frame temporal features for saliency inference. Here, it reveals the other issue, which is how to effectively integrate the spatial and temporal features in video saliency detectors. It is insufficient to separately exploit them to detect the salient objects, which may lead to the failure and leak detection in VSOD. 

Before the wide employment of deep models, the heuristic models \cite{wang2015consistent} are mainstream solutions for VSOD. Relying on the hand-crafted features and optimization model, these approaches cannot guarantee the performance and runtime of the saliency detection. With the introduction of deep learning, the performance of VSOD has been significantly improved. However, as for some complex video scenes (e.g., low resolution, motion blur, scale change, etc), the early deep models \cite{8047320,8419765} are limited to extract robust spatial and temporal features by the straightforward fully convolutional network (FCN). To obtain more robust and reliable features, especially temporal features, optical flow and long-short term memory (LSTM) are exploited in this field. Although the recent approach adopts optical flow and LSTM modules to obtain temporal features, they still effectively integrate the spatiotemporal features by naive fusion strategies \cite{tang2020video,song2018pyramid,Fan_2019_CVPR} such as addition or concatenation. 

Such fusion strategies though greatly simply the resulting framework, they are insufficient to discover the correlation of features from different domains, thus generate more robust fusing features for saliency prediction in video scenes. In order to solve the above issue, we propose a new two-stream encoder-decoder architecture, called adaptive local-global refinement network (LGRN), for video salient object detection. First, to alleviate the feature redundancy caused by the simple fusion strategies, we design a local refinement module (LRM) and a global refinement module (GRM), which can fully refine the multi-level features in both local and global scopes. Further, a weighting strategy based on GCN is devised to adaptively emphasize important information of different types of features. 
%
The main contributions of this paper are summarized as

\begin{itemize}
	\item We design a local-global refinement network to refine the simply fused features in a hierarchical manner. Different from existing methods merely adopting a simple fusion strategy, our refinement scheme can fully exploit the local and global dependence of multi-level features.   
	\item An adaptive weighting strategy based on graph convolutional neural network (GCN) is devised to further exploit the correlations of different types of features. Through emphasizing the effective information and suppressing the useless one, our framework is driven to learn more discriminative representation for VSOD.  
	\item The experiments on the widely used datasets demonstrate that the proposed approach achieves competitive performance against the state-of-the-arts. 
\end{itemize}

\section{Related Work}

\textbf{Image and video salient object detection} Early image saliency detectors mainly utilize hand-crafted features and heuristic models. As the limitation of hand-crafted features and low efficiency of heuristics, these approaches are not able to handle complex scenes and be applicable for practical systems. With the wide deployment of convolutional neural networks (CNNs), the performance of image saliency has obtained significant improvement. The original CNN-based saliency detectors are to replace hand-crafted features with deep features \cite{li2016visual,wang2015deep}, but the heuristic models are retained. Therefore, this kind of method is still low efficiency. After that, the fully convolutional network (FCN) is introduced in this field. Due to the high efficiency and performance of FCN, many approaches \cite{wei2020f3net,deng2018r3net,wu2019cascaded,chen2020reverse} exploits it to obtain more robust deep features for saliency prediction. 

The development of video saliency is very similar to image saliency. Before the rise of deep learning, heuristic models are also widely used in video saliency. The difference is that video saliency detectors need to explore the employment of temporal information. Conventional video heuristics usually exploits the optical flow to jointly optimize the models. After the introduction of FCN, the end-to-end framework has become the principal method. Video saliency models also exploit the encoder-decoder model, but some extra sequential modules \cite{song2018pyramid,Li_2019_ICCV,li2018flow,Fan_2019_CVPR} are inserted into their frameworks to extract temporal features. For example, PDB \cite{song2018pyramid} proposes a pyramid dilated convolutional LSTM module to extract sequential features in the deep models. MGA \cite{Li_2019_ICCV} proposes an optical flow attention module to encode the multi-level deep features in the FCN-based framework.

\noindent \textbf{Graph neural network}. Recently, graph neural network (GNN) has been applied for a variety of visual tasks, such as vehicle re-identification \cite{mm/LiuLZY020}, visual question answering \cite{gao2020multi}, human pose regression \cite{zhao2019semantic}. Generally, GNN can be divided into two categories: spectral methods and non-spectral methods. The former one conducts the convolution transform in the Fourier domain, whose representation is the graph convolutional network (GCN). In \cite{mm/LiuLZY020}, PCRNet extract the part-level features from vehicle parsing and adopts GCN to capture the cross-part relationship for vehicle re-identification. GCAGC \cite{zhang2020adaptive} proposes an adaptive GCN, which is used to discover the non-local and long-range correspondence in an image group for co-saliency detection. The latter one is non-spectral methods, also called spatial-based methods. Graph attention network (GAT) belongs to this scope and is widely used in many visual tasks. In \cite{luo2020cascade}, GAT is used to explore the correlation between RGB image and depth image for RGB-D salient object detection. In \cite{lu2020video}, Lu et al. propose a neat and fast graph memory network, which modifies the message passing mechanism, and allow the network to capture motion cues for video object segmentation. 

\begin{figure*}[t]
	\centering
	\includegraphics[width=1\textwidth]{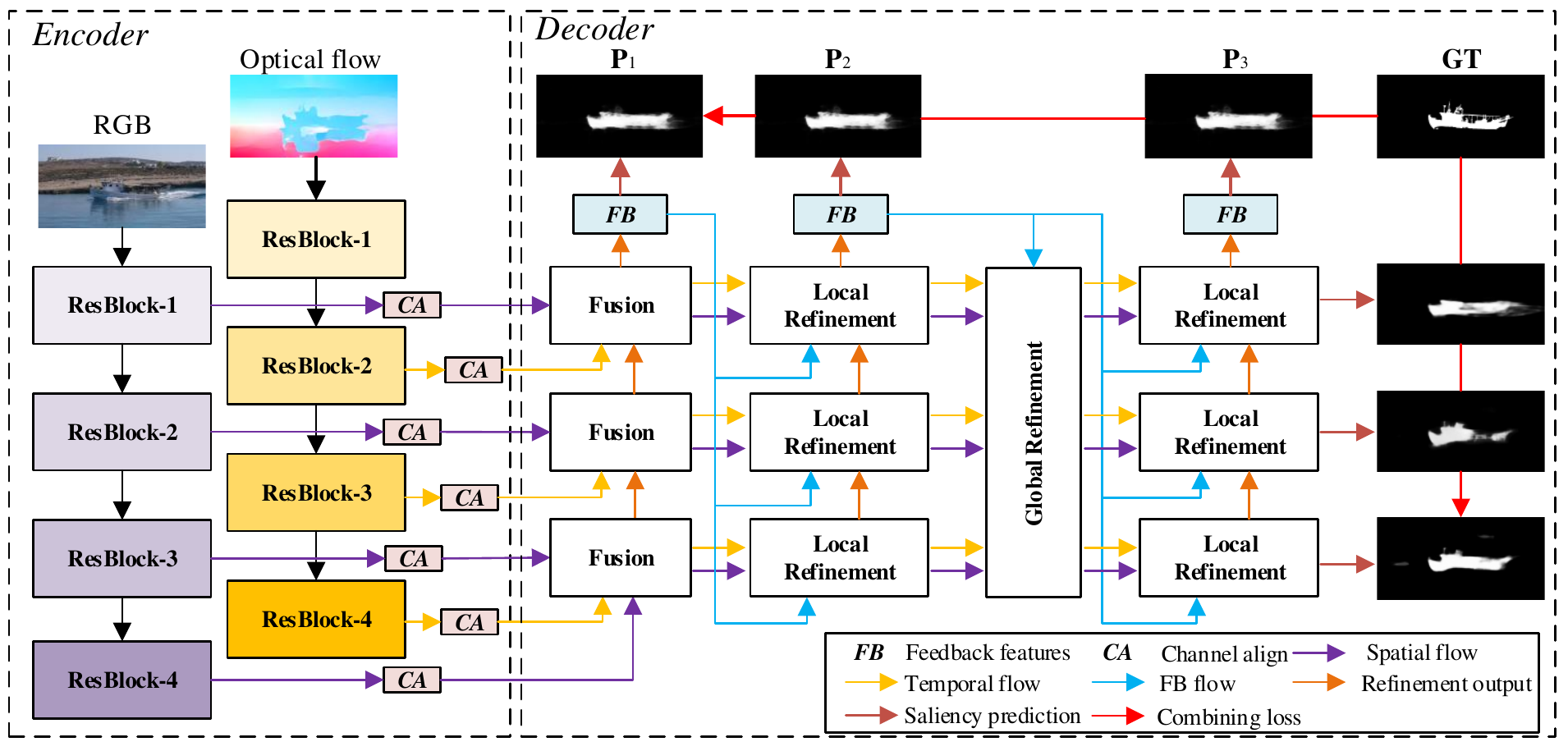}
	\caption{The framework of our proposed LGRN. The encoder is regarded as a feature extractor to generate multi-level spatial and temporal features. The decoder consists of two local refinement modules and a global refinement module to exploit the correlations among multi-level features with different scopes.}
	\label{framework}
	\vspace{-10px}
\end{figure*}


\section{Proposed method}
The whole framework of our proposed LGRN is illustrated in Fig.~\ref{framework}. We can see that LGRN exhibits as a two-stream encoder-decoder architecture, which mainly includes two components, i.e., the encoder and the decoder. 

The encoder consists of two feature extractors to generate the multi-level features from the RGB stream and the optical flow stream. The primary contribution of our work lies in the design of the decoder.  On the decoder side, we first propose to fuse the features from different sources in a hierarchical manner. Further, a local-global refinement architecture is operated to refine the simply fused features, and finally generate the predicted saliency map. Different from existing methods merely adopting a simple fusion strategy, our hierarchical refinement scheme can fully exploit the local and global dependence of multi-level features, making the proposed LGRN learn more discriminative feature representation for VSOD.  

\subsection{Generation and fusion of multi-level features}
On the encoder side, we use two backbone networks for the feature extraction, then to generate multi-level features from each RGB video frame and the corresponding optical flow map. For simplicity, denote $I^s\in \mathbb{R}^{w\times h\times 3}$ as a generic RGB frame of a video sequence, with the optical flow $I^t$. Then the multi-level features from spatial and temporal domains can be extracted as follows:
\begin{equation}
	\{I^s_l\}_{l=1}^{L_s} = BBN_s(I^s), ~~	\{I^t_l\}_{l=1}^{L_t} = BBN_t(I^t),
\end{equation} where $BBN_s$ and $BBN_t$ represent the backbone networks for the spatial domain and the temporal domain, respectively. $I^s_l \in \mathbb{R}^{w\times h\times c_s}$ and $I^t_l \in \mathbb{R}^{w\times h\times c_t}$ denote the features extracted from the $l$-th block, where $c_s$ and $c_t$ are the channel sizes of the $l$-th layer. $L_s$ and $L_t$ serves as the number of the convolutional blocks of two networks. 

Note that for a specific layer $l$, the number of channels could be different for spatial stream and temporal stream. Even for the same stream, the number of channels could also be different across different layers. To handle this issue, we adopt a channel align technique to adjust the channel size. Mathematically, 
\begin{equation}
	\label{extraction}
	\begin{aligned}
		f^{s}_{l} = CA(I^{s}_{l}), ~~f^{t}_{l} = CA(I^{t}_{l}),
	\end{aligned}
\end{equation}  where $CA$ is a convolutional operation with batchnorm and Relu activation function. By resorting to the channel align technique, all the features have the  and $f^{t}_{l} \in R^{w\times h \times c_0}$ are the aligned feature maps. Through this operation, we can obtain a set of multi-level feature maps $\{f^{s}_{l}\}_{l=1}^{L_s}$ and $\{f^{t}_{l}\}_{l=1}^{L_t}$ with same dimension $(w\times h\times c_o)$.

Upon having the features from spatial and temporal domains, we fuse them in a hierarchical manner as shown in Fig. \ref{framework}.  For the fusion in level $l$, the multi-level features from different domains are integrated by a element-wise multiplication, whose process can be written as below:

\begin{equation}
	\label{refine1}
	\begin{aligned}
		f^{s}_{l}&=f^{s}_{l}+ Conv(f^{s}_{l} * f^{s}_{l-1} * f^{t}_{l}) \\
		f^{s}_{l-1}&=f^{s}_{l-1}+ Conv(f^{s}_{l} * f^{s}_{l-1} * f^{t}_{l}) \\
		f^{t}_{l}&=f^{t}_{l}+ Conv(f^{s}_{l} * f^{s}_{l-1} * f^{t}_{l}) \\
	\end{aligned}
\end{equation} where  $Conv(\cdot)$ is a convolution with Batchnorm and Relu. $f^{s}_{l}$ and $f^{t}_{l}$ are the corresponding spatial and temporal features.

Note that since the RGB frame and optical flow map are highly correlated, simply fusing them with multiplication can effectively suppress the background noise and extract the overlapping salient
region against the other naive strategies (e.g., addition or concatenation.), but it will lose some boundary information, thus cannot highlight the whole salient objects and severely degrade the performance of the resulting saliency detector \cite{8419765}. To address this problem, in this paper, we propose a local-global refinement technique to exploit the local dependence and the global dependence of multi-level features, thus refine the simply fused features with different scopes. The architecture of our local-global refinement module is shown in Fig. \ref{framework}.

\subsection{Local refinement}
The goal of the local refinement module (LRM) is to refine the fused features hierarchically. As shown in Fig. \ref{framework}, our local refinement architecture consists of a set of local refinement blocks (LRB), where each block aims to exploit the correlations among features in adjacent layers. 

To make full use of the intermediate detection results, we adopt the feedback mechanism as marked by blue lines in Fig. \ref{framework}.  As will be shown in the experimental stage, the feedback mechanism plays an important role for the LRB to select useful and reliable information. Furthermore, we propose an adaptive weighting mechanism based on GCN to fully exploit the correlations of the input features. Our weighting methodology can enforce the LRB to select effective information and suppress the useless one, thus driving the resulting network to learn more discriminative feature representation.

\begin{figure}[t]
	\centering
	\includegraphics[width=0.5\textwidth]{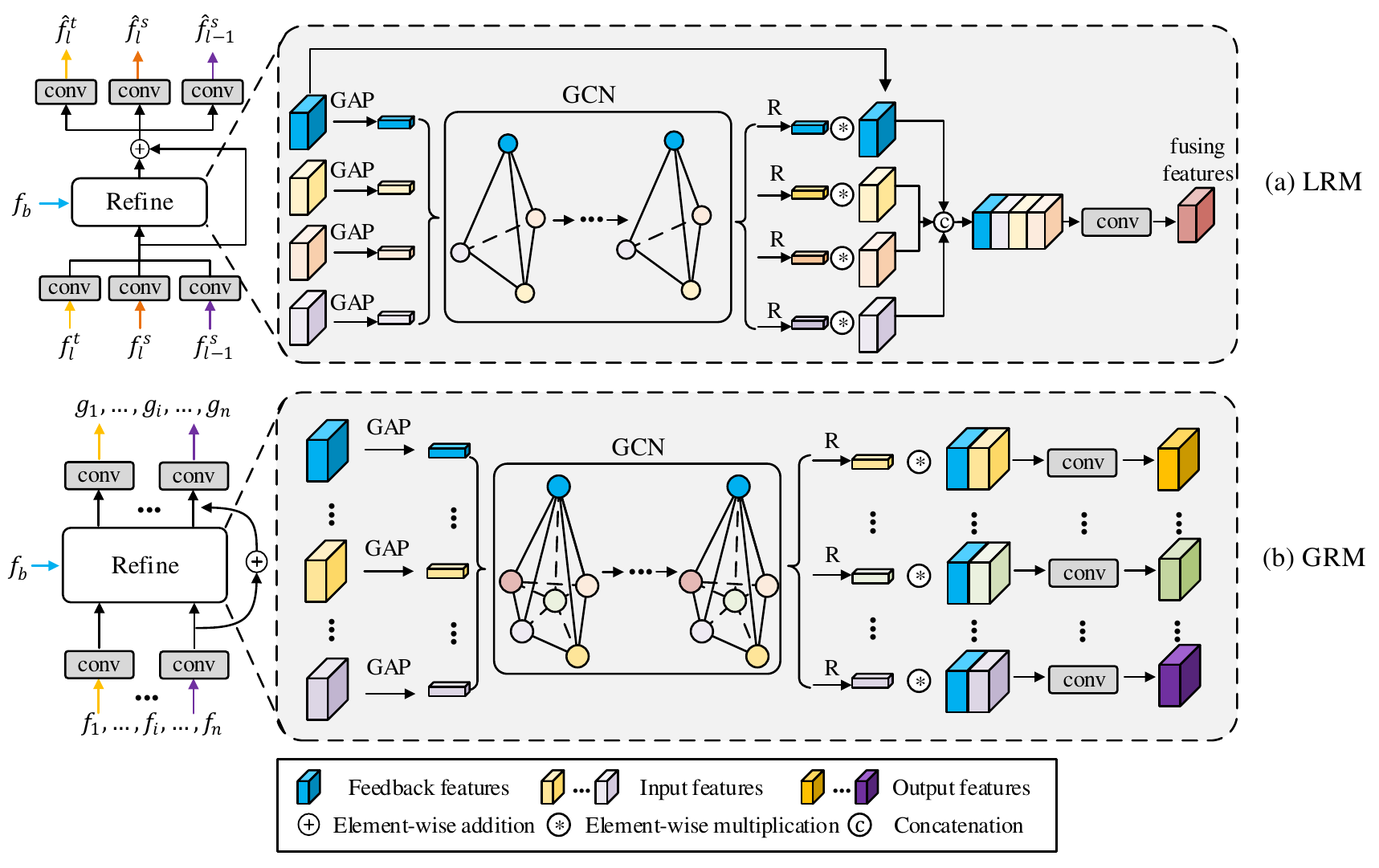}
	\caption{The adaptive local-global refinement. (a) The pipeline of local refinement module (LRM). (b) The pipeline of global refinement module (GRM).}
	\label{LRB}
	\vspace{-10px}
\end{figure}

The framework of our LRB is illustrated in Fig. \ref{LRB} (a). The LRB takes input the high-level spatial features $f^{s}_{l}$, the low-level features $f^{s}_{l-1}$, the temporal features $f^{t}_{l}$ and the feedback features $f_{b}$.  For simplicity, we use $F_l= \{f_1, f_2, f_3, f_b\}$ as the four inputs. To fully exploit the correlations of these four features, we propose to construct their relationships as a graph. Denote the graph as $G_l=\{V_l, E_l\}$, where $V_1 = \{v_1, v_2, v_3, v_b\}$ represents the four vertices of $G_l$ and  $E_l$ contains all the edges. The attribute of $v_i$ is vector, which is obtained by applying a group of convolutional operations (kernel size 3$\times$3) and global average pooling (GAP) (size 1$\times$1) to the input feature $f_i$ \footnote{Hereafter, we will interchangeably use $v_i$ to represent a node or its attribute.}. Furthermore, to characterize how strongly that two features are correlated, we build a weighted adjacency matrix $A_l$ according to their cosine similarities. Then the element of $A_l$ can be computed as: 
\begin{equation}\label{eq:adjacent}
	a_{i,j} =  cos(v_i, v_j), ~v_i, v_j \in V_l. 
\end{equation}
Upon constructing the graph, we adopt GCN to learn adaptive weights through information propagation from each node to its neighbors. Let $X^{(m)}$ be the representation of the node $v_i$ in the $m$-th layer of GCN, and we set $X^{(0)} = V_l$. 
According to \cite{kipf2017Arxiv}, we design the GCN layer as:
\begin{equation}
	X^{(m)} = \sigma(D_l^{-\frac{1}{2}}A_lD_l^{-\frac{1}{2}}X^{(m-1)}W_l^{(m)}),\\
\end{equation} where $\sigma$ is a certain activation function, $D_l$ is the diagonal node degree matrix of $A_l$, and $W_l^{(m)}$ is a set of associated learnable parameters of the $m$-th layer. 

Assume that there are $M_l$ GCN layers for each LRB. The weights indicate the importance of the input features are then generated by $r = R(X^{(M)})$, where $R(\cdot)$ is a set of fully-connected layers with sigmoid activation. Finally, we propose to refine the input features by consecutively employing the following three operations:
\begin{equation}
	\label{merge}
	\begin{aligned}
		\tilde{f}_i &= r_i * f_i, f_i \in F_l\\ 
		\tilde{F} & = Cat(\tilde{f}_1, \tilde{f}_2, ..., \tilde{f}_i) \\
		\hat{F}_l &= G_l(\tilde{F}, W_l), 
	\end{aligned}
\end{equation}  where $\hat{F}_l$ represents the refined feature maps, $Cat$ is the concatenation operation and $G_l(\cdot)$ denotes a $1\times 1$ convolution layer with the parameter $W_l$. 

By resorting to the designed LRM, the input features of adjacent levels are refined. The adaptive weighting strategy fully exploits the relationships among different types of features, and enforce LRM to emphasize important information and suppress useless one. 



\subsection{Global refinement}
The local refinement merely though is effective to refine features in adjacent feature levels, it fails to exploit the correlations among features in the global scope. To handle this issue, we propose a global refinement module (GRM) to further refine the features globally. 

The architecture of GRM is shown in Fig.~\ref{LRB} (b). The GRM takes the input of a collection of multi-level features generated from LRMs and feedback features. For the sake of simplicity, we denote the input features as $F_g = \{\hat{f}_1, ... \hat{f}_n, \hat{f}_b\}$. To exploit the correlations among features in the global scope, similar to LRB, we construct a graph $G_g = \{V_g, E_g\}$ to represent their mutual relationships. The notation $V_g = \{\hat{v}_1, ..., \hat{v}_n, \hat{v}_b\}$ and $E_g$ denote the vertices and edges of $G_g$. The attribute of $v_i$ is computed through a set of convolutional layers and a GAP operator. Then we propose to learn the adaptive weights by a ground of GCN layers, where the message propagation process can be formulated as:
\begin{equation}
	\label{gcn}
	\begin{aligned}
		\hat{X}^{(m)} &= \sigma(D_g^{-\frac{1}{2}}A_gD_g^{-\frac{1}{2}}X^{(m-1)}W_g^{(m)}), m = 1,..,M \\
		\hat{r} &= R(\hat{X}^{(M)} ).
	\end{aligned}
\end{equation} Here $A_g$ is the weighted adjacency matrix of $G_g$ to encode the strength of correlation between features, which can be similarly constructed as (\ref{eq:adjacent}). $D_g$ is the degree matrix of $A_g$, and $W_g$ contains the associated parameters. 
Finally, GRM refine the multi-level features using global information as 
\begin{equation}
	\label{grm}
	\begin{aligned}
		\tilde{f}_i &= \hat{r}_i * Cat(\hat{f}_i, \hat{f}_b),\\ 
		g_i &= G_i(\tilde{f}_i, W_i), i \in 1, ..., n, 
	\end{aligned}
\end{equation} where $G_i(\cdot)$ is a 1$\times$1 convolution and $g_i$ represents the refined version of $\hat{f}_i$. Note that different from $\hat{f}_i$, which is only refined by LRM using local information, the feature $g_i$ refined by GRM can capture more global information. As shown in Fig.~\ref{framework} the LRMs can reuse the global information propagated from the GRM, which can effectively integrate the local and global information and further discover their correlation for final saliency prediction.

Different from the previous works merely using the naive fusion strategies, such as addition or concatenation, the proposed adaptive local-global refinement can discover the correlations among features from different domains. The refinement technique allows our method to automatically emphasize the effective information and suppressing the useless one, thus to learn more discriminative representation.

\begin{table*}[]
	\centering
	\begin{tabular}{c|ccc|ccc|ccc}
		\hline
		\multirow{2}{*}{Methods}   & \multicolumn{3}{c|}{DAVIS}                             & \multicolumn{3}{c|}{ViSal}                             & \multicolumn{3}{c}{DAVSOD}                            \\
		& $F_\beta$$\uparrow$ & S$\uparrow$    & MAE$\downarrow$ & $F_\beta$$\uparrow$ & S$\uparrow$    & MAE$\downarrow$ & $F_\beta$$\uparrow$ & S$\uparrow$    & MAE$\downarrow$ \\ \hline
		R3Net \cite{deng2018r3net} & 0.859               & 0.881          & 0.027           & 0.937               & 0.910          & 0.024           & 0.605               & 0.753          & 0.085           \\
		CPD \cite{wu2019cascaded}  & 0.838               & 0.841          & 0.035           & 0.929               & 0.934          & 0.022           & 0.608               & 0.704          & 0.079           \\
		RAS \cite{chen2020reverse} & 0.860               & 0.873          & 0.032           & 0.941               & 0.934          & 0.022           & 0.613               & 0.727          & 0.079           \\
		F3Net \cite{wei2020f3net}  & 0.866               & 0.868          & 0.035           & 0.935               & 0.926          & 0.026           & 0.629               & 0.756          & 0.085           \\ \hline
		SCOM \cite{Chen2018TIP}    & 0.783               & 0.832          & 0.048           & 0.831               & 0.762          & 0.122           & 0.464               & 0.599          & 0.220           \\
		SCNN \cite{8419765}        & 0.714               & 0.783          & 0.064           & 0.831               & 0.841          & 0.071           & 0.532               & 0.674          & 0.128           \\
		FGRNE \cite{li2018flow}    & 0.783               & 0.838          & 0.043           & 0.848               & 0.861          & 0.045           & 0.573               & 0.693          & 0.098           \\
		PDB \cite{song2018pyramid} & 0.855               & 0.882          & 0.028           & 0.888               & 0.907          & 0.032           & 0.572               & 0.698          & 0.116           \\
		SSAV \cite{Fan_2019_CVPR}  & 0.861               & 0.893          & 0.028           & 0.939               & 0.943          & 0.020           & 0.603               & 0.724          & 0.098           \\
		MGA \cite{Li_2019_ICCV}    & 0.902               & 0.913          & 0.022           & 0.947               & 0.944          & 0.015           & 0.646               & 0.734          & 0.073           \\
		TEN \cite{ren2020tenet}    & 0.893               & 0.905          &\textbf{0.017}          & 0.949               & 0.946          & 0.014           & 0.684               & 0.757          & 0.067           \\
		Ours                       & \textbf{0.920}      & \textbf{0.923} & \textbf{0.017}  & \textbf{0.954}      & \textbf{0.951} & \textbf{0.013}  & \textbf{0.699}      & \textbf{0.765} & \textbf{0.063}  \\ \hline
	\end{tabular}
	\caption{Quantitative comparison results by different ISOD approaches and VSOD approaches on DAVIS, ViSal, DAVSOD datasets.}
	\vspace{-5px}
	\label{compare_all}
\end{table*}

\begin{figure*}[t]
	\centering
	\includegraphics[width=1\textwidth]{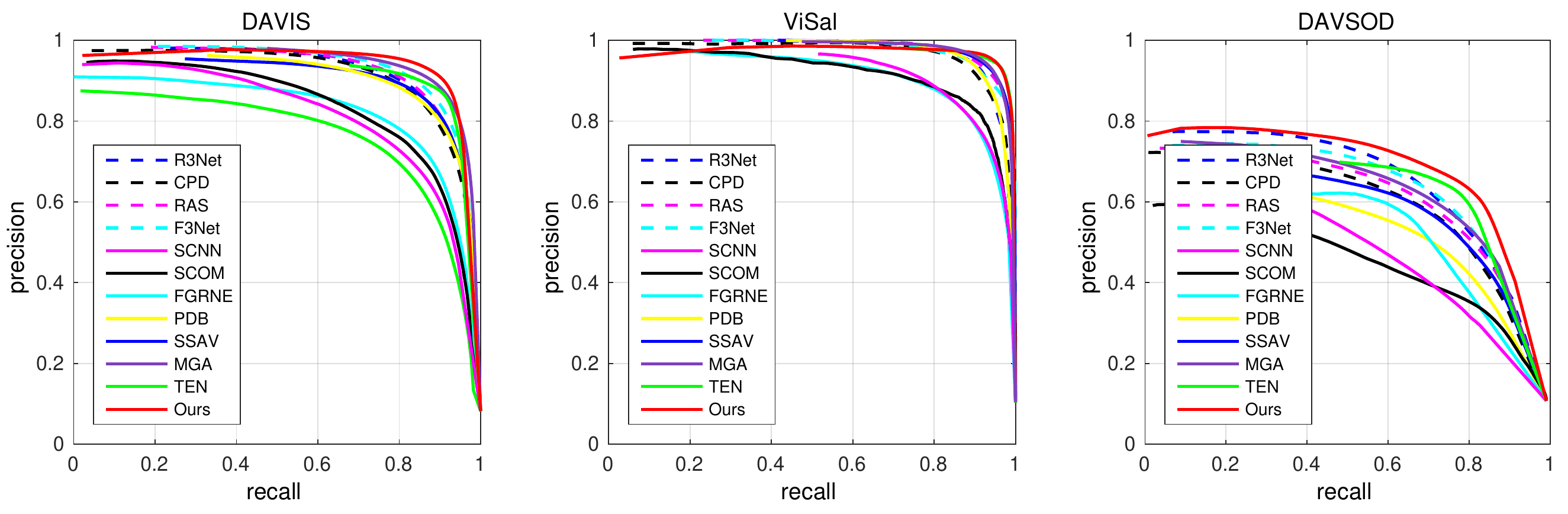}
	\vspace{-20px}
	\caption{Comparison of precision-recall curves between ISOD methods and VSOD methods on DAVIS, ViSal and DAVSOD dataset.}
	\vspace{-5px}
	\label{PRC}
\end{figure*}

\subsection{Loss function}
In order to train our framework, we first consider the widely employed binary cross-entropy loss (BCE) function. The BCE measures the distance between the predicted saliency map and ground truth, which is given by: 
\begin{equation}
	\label{loss1}
	\begin{aligned}
		L_{bce}(S, GT)  &= -\sum_{i=1}^{w*h}[g_{i}logP(s_i) + (1 - g_i)log(1 - s_i)],
	\end{aligned}
\end{equation} where $s_i \in S$ and $g_i \in GT$ represent the probability of the predicted saliency map $S$ and the label of ground truth $GT$ at the location $i$, respectively. 

Except for BCE loss, in this work, we also adopt a combination loss to train our framework, which is composed of BCE, IoU loss and Focal loss. 
The IoU loss $L_{IoU}$ is a widely used in segmentation task, which calculates the similarity between the salient region and the ground truth. IoU loss is defined as

\begin{equation}
	\label{loss2}
	\begin{aligned}
		L_{IoU}(S, GT)  &= 1-\frac{\sum_{i=1}^{w*h}s_i * g_i}{\sum_{i=1}^{w*h}(s_i + g_i - s_i * g_i)},
	\end{aligned}
\end{equation}

The focal loss $L_{foc}$ is designed to mitigate the problem of class unbalance, which can be formulated as

\begin{equation}
	\label{loss3}
	L_{foc}(S, GT)=
	\begin{cases}
		-\alpha(1-s_i )^{\gamma} logs_i & g_{i}=0\\
		-(1-\alpha)s_i^{\gamma} log(1-s_i) & g_i \neq 0
	\end{cases}
\end{equation} where $\alpha$ is a balance factor, $\gamma$ is a parameter used to reduce the loss for well-classified samples and emphasize on the misclassiﬁed ones. 
Finally, the combination loss is defined as 

\begin{equation}
	\label{loss2}
	\begin{aligned}
		L  &= L_{bce} + L_{IoU} + L_{foc},
	\end{aligned}
\end{equation}

\begin{figure*}[t]
	\centering
	\includegraphics[width=1\textwidth]{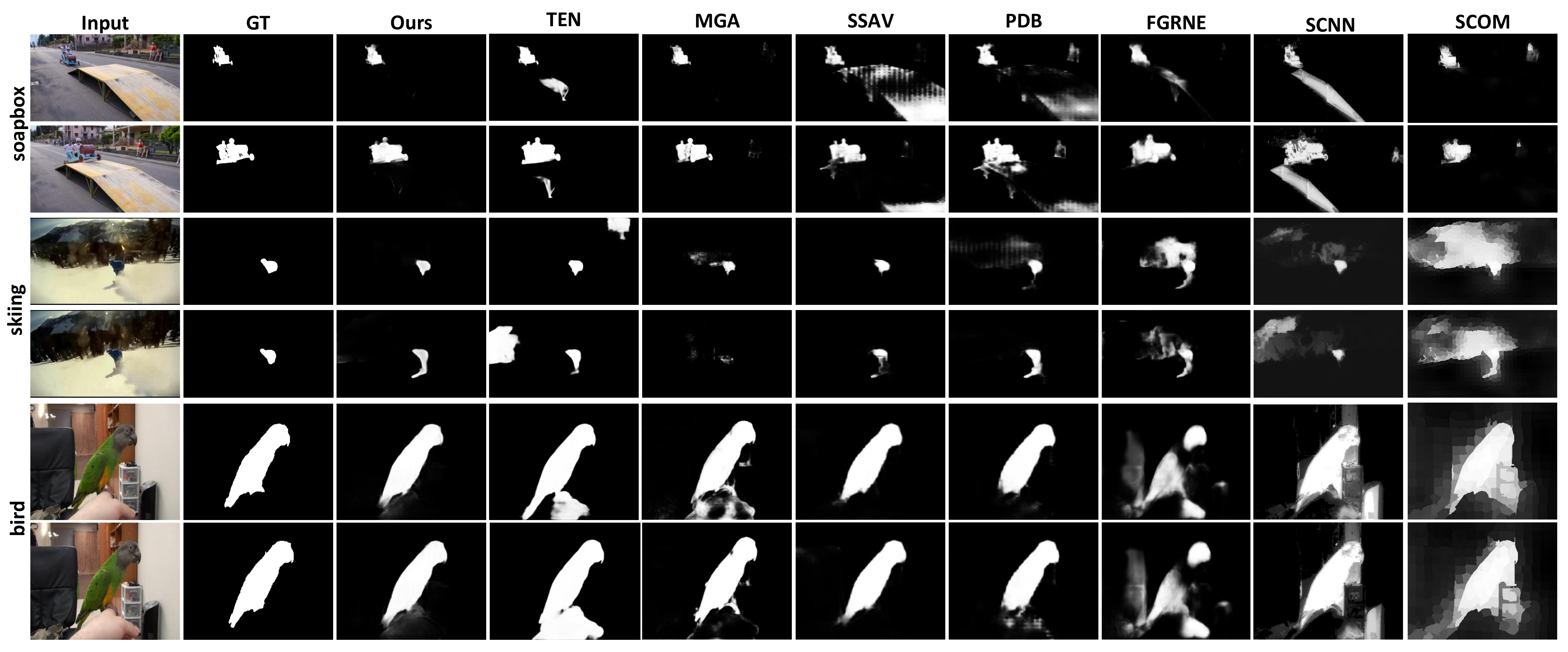}
	\caption{Saliency maps generated by the state-of-the-art methods and the proposed one. Notice that the saliency maps by the proposed approach achieve competitive performance in different video scenes. }
	\label{show}
\end{figure*}

\section{Experiments}

\subsection{Experimental settings}
\textbf{Datasets.} In this paper, we evaluate the proposed network on three datasets, including DAVIS \cite{perazzi2016benchmark}, ViSal \cite{wang2015consistent}, and DAVSOD \cite{Fan_2019_CVPR}. Among them, DAVIS contains 50 video sequences, which are divided into a training set (30 videos) and a testing set (20 videos). ViSal includes 17 video clips and 193 frames in total. DAVSOD is the most difficult dataset of VSOD in recent datasets. It contains 61 video clips in the training set and 35 video clips in the testing set. In our experiments, we use the training set of DAVIS and DAVSOD for network training, while the remains are used for testing. 

\noindent \textbf{Evaluation criteria.} As for the evaluation criteria, we follow the setting in MGA \cite{Li_2019_ICCV}. Three different evaluation methods are adopted for the comparison between the proposed one and the others. They are maximum F-measure (F$_\beta$), S-measure (S$_m$) and mean average error (MAE). F$_\beta$ measures the harmonic mean of precision and recall between the predicted saliency map and ground truth. It can be fomulared as below:
\begin{equation}
	\label{f_measure}
	\begin{aligned}
		F_{\beta} = \frac{(1 + \beta^2) \cdot Precision \cdot Recall}{\beta^2 \cdot Precision + Recall},
	\end{aligned}
\end{equation} where $\beta^2$ is set to 0.3, and $Precision$ and $Recall$ can be obtained by computing the average value of saliency maps. S-measure considers both region and object structural similarity.

S-measure considers both region and object structural similarity, whose defination can be wriitten as:

\begin{equation}
	\label{s_measure}
	\begin{aligned}
		S= \mu * S_o + (1 - \mu) * S_r,
	\end{aligned} 
\end{equation} where $S_o$ and $S_r$ are the region-aware structural similarity and object-aware structural similarity, respectively. $\mu$ is balance factor, which is set to 0.5.

MAE calculates the mean square errors between the saliency maps and the corresponding ground truths. It can be demonstrated as follow:

\begin{equation}
	\label{f_measure}
	\begin{aligned}
		MAE = \frac{1}{|\mathcal{S}|} \sum_i |\mathcal{S}(p_i) - \mathcal{G}(p_i)|,
	\end{aligned}
\end{equation} where $p_i$ represents pixels in a frame; $\mathcal{S}$ and $\mathcal{G}$ are the saliency map and the corresponding ground truth, respectively.  Additionally, we also introduce the Precision-recall (PR) curve for the evaluation in this paper. It shows the relationship between precision (positive predictive value) and recall (sensitivity) for every possible cut-off.

\noindent \textbf{Implementation details.} We implement the proposed approach by PyTorch. During the training, except for the training set of DAVIS and DAVSOD, we also use DUT-TR \cite{wang2017learning} to train the spatial stream. To obtain optical flow maps, we utilize FlowNet2.0 \cite{8099662} as the extractor. The backbones of the spatial and temporal stream are ResNet-50 and ResNet-34, respectively. As for the network hyperparameters, we adopt SGD as the optimizer. The initial learning rate is 0.005, which follows the ``poly" adjustment policy. The momentum is 0.925 and weight decay is 0.0005. For data augmentation, we use random cropping (crop size: 380 $\times$ 380), random rotation (10 degrees) and random horizontal flipping\footnote{For the specific network parameters, please check our released code.}. Our hardware is an Nvidia Geforce TITAN X GPU.

\subsection{Comparison with the state-of-the-arts}

In this paper, we totally compare 11 approaches, which include , R3Net \cite{deng2018r3net}, CPD \cite{wu2019cascaded}, RAS \cite{chen2020reverse}, F3Net \cite{wei2020f3net}, 4 ISOD approaches and SCOM \cite{Chen2018TIP}, SCNN \cite{8419765}, FGRNE \cite{li2018flow}, PDB \cite{song2018pyramid}, SSAV \cite{Fan_2019_CVPR}, MGA \cite{Li_2019_ICCV}, TEN \cite{ren2020tenet}, 8 VSOD approaches.

For the ISOD approaches, we finetune their deep models with the video frames from DAVIS and DAVSOD datasets. Table.~\ref{compare_all} shows the quantitative comparison results. Notice that the proposed approach achieves competitive performance. In DAVIS and ViSal, the video scenes are relatively easy. Recent VSOD approaches can obtain high performance. As our method explores the correlation between spatial and temporal features, we can further improve the performance. DAVSOD dataset contains many complex and multiplies object scenes, but our method still outperforms the others in all of the criteria without any post-processing and online learning strategies like TEN \cite{ren2020tenet}. Moreover, the proposed method outperforms the second best by 2.2\%, 1.1\% and  6.0\% in terms of F$_\beta$, S-measure and MAE, respectively.

As shown in Fig.\ref{PRC}, the curve of our method outperforms the others at the right-top corner. It demonstrates that the proposed one simultaneously hold outstanding precision and recall. Due to performance saturation, our method still obtain moderate improvement. In DAVIS and DAVSOD, the proposed one obviously increases in term of precision and recall axis against the others.  

Fig.~\ref{show} shows the visual comparison between the proposed one and other approaches. Apparently, our saliency maps have clearer boundaries and eliminate much background noise. For example, in \textit{soapbox} video sequences, the salient object suffers from the interference of high contrast background regions. Our method can overcome this issue and highlight the whole salient object against the other approaches. Meanwhile, as for other complex scenarios, like motion blur variation (\textit{skiing}), we also can capture robust spatiotemporal cues and accurately detect the salient objects, but the others still exist some background noise in their saliency maps. For example, the saliency maps from TEN pop out big non-salient regions, which decrease the performance in precision and recall.


\subsection{Ablation studies}
\begin{table}[t]
	\centering
	\begin{tabular}{cccc|ccc}
		\multirow{2}{*}{BCE} & \multirow{2}{*}{CL} & \multirow{2}{*}{LRM} & \multirow{2}{*}{GRM} & \multicolumn{3}{c}{ViSal}                             \\
		&                     &                      &                      & $F_\beta$ $\uparrow$ & S $\uparrow$ & MAE $\downarrow$ \\ \hline
		$\checkmark$         &                     &                      &                      &   0.942          &   0.936              & 0.025            \\
		& $\checkmark$        &                      &                      & 0.943               &    0.938          & 0.016            \\
		& $\checkmark$        & $\checkmark$         &                      & 0.949                &  0.944            & 0.014            \\
		& $\checkmark$        & $\checkmark$         & $\checkmark$         & \textbf{0.954}                & \textbf{0.951}      & \textbf{0.013}            \\ \hline
	\end{tabular}
	\caption{Ablation study on ViSal. BCE represents that the model is trained only by BCE loss function and does not exploit the proposed modules. CL represents the combining loss function. LFM and GRM are the proposed modules in our architecture.}
	\label{diff_compare}
\end{table}

\begin{table}[]
	\centering

	\begin{tabular}{ccc|ccc}
		\multirow{2}{*}{FC} & \multirow{2}{*}{GCN} & \multirow{2}{*}{FB} & \multicolumn{3}{c}{ViSal}                               \\
		&                      &                    & $F_\beta$ $\uparrow$ & S $\uparrow$   & MAE $\downarrow$ \\ \hline
		&                      &       $\checkmark$ &0.939                & 0.932          & 0.021          \\
		& $\checkmark$         &                     & 0.941                & 0.935          & 0.017            \\
		${\checkmark}$ &                      & $\checkmark$        & 0.947                & 0.938          & 0.015            \\
		& $\checkmark$         & $\checkmark$        & \textbf{0.954}       & \textbf{0.951} & \textbf{0.013}   \\ \hline
	\end{tabular}
	\caption{The effectiveness of the channel weighting, GCN and feedback features.}
	\vspace{-10px}
	\label{op_compare}
\end{table}

In order to validate the effectiveness of the combing loss function, LFM and GFM, we conduct several experiments on DAVIS and ViSal datasets. Table.~\ref{diff_compare} shows the ablation study for different modules. We design a baseline for comparison. It is an encoder-decoder architecture, but it does not introduce the proposed modules. Its feature fusion of adjacent residual blocks is using element-wise multiplication. Moreover, the network is trained only by the BCE loss function. With the gradual introduction of the proposed modules, the performance is steadily improved. It demonstrates that all of the proposed modules are effective.

Additionally, we design an experiment to validate the effectiveness of the GCN and feedback features. Firstly, we directly remove the channel weighting module, namely GCN and only use feedback features for saliency prediction. Secondly, we remove the feedback features in the proposed modules and then train the network. Thirdly, in order to validate the effectiveness of GCN, we use fully-connected operations instead of GCN to learn the feature weights. Finally, all of components are used to train the network. The results are shown in Table.~\ref{op_compare}. As we can see, the feedback features and GCN is the important components in the proposed modules. The former contains reliable high-level semantic information, which can be used to guide the network training. The latter can build the relationship between the multi-level features from different domains and discover their complementary dependence for saliency prediction.

\subsection{Runtime analysis}

\begin{table}
	\centering
	\begin{tabular}{|c|c|c|c|c|}
		\hline
		Method  & Ours  & TEN  & MGA  & SSAV  \\ \hline
		Time(s) & 0.04  & 0.06 & 0.07 & 0.05  \\ \hline
		Method  & PDB   & FGRN & SCNN & SCOM  \\ \hline
		Time(s) & 0.05  & 0.09 & 1.79 & 2.11  \\ \hline
		Method  & F3Net & RAS  & CPD  & R3Net \\ \hline
		Time(s) & 0.01  & 0.01 & 0.02 & 0.03  \\ \hline
	\end{tabular}
	\caption{Average runtime comparison between the proposed approach and the other salient object detection approaches.}
	\label{runtime}
	\vspace{-10px}
\end{table}

In Table.~\ref{runtime}, we report the runtime comparison between the proposed methods and the others. To be consistent with the other works like \cite{Li_2019_ICCV} and \cite{li2018flow}, the reported runtimes do not contain the cost of the optical flow processing. As shown in Table.~\ref{runtime}, the the image saliency methods, such as F3Net \cite{wei2020f3net} and RAS \cite{chen2020reverse}, are faster than the video saliency methods because their network structures are single-stream. Besides, they cannot use the sequential modules, like RNN or LSTM. Compared with them, our method is slower but still still more efficient than the previous video saliency methods, like MGA \cite{Li_2019_ICCV}  and TEN \cite{ren2020tenet}. 

For the proposed network, we also measure the runtime of each component. We mainly have four components. the backbone costs 0.013s, the preliminary multi-level fusion modules cost 0.006s, the local refinement modules cost 0.007s, the global refinement module costs 0.002s. The total runtime is 0.014 + 0.006 + 0.009*2 + 0.004 = 0.042. Through the analysis of each component, we can note that the cost of backbone account for 1/3 of the entire runtime. If we choose deeper network as the backbone like ResNet101, the runtime will furthre increase. Therefore, we choose ResNet50 as the backbone, which not only ensures the competitive performance, but also achieves high runtime efficiency. Additionally, although the proposed local and glocal refinement modules look complicated, the runtime is not slow. For example, compared with the preliminary multi-level fusion modules by using naive fusion strategy, the proposed local refinement modules are slight slower. 

\section{Conclusion}

In this paper, we firstly propose a local-global refinement network for video salient object detection, which is able to refine the multi-level features in a hierarchical manner. Secondly, based on GCN, an adaptive weighting strategy is proposed to discover the correlations of different domains of features. Through highlight the effective information and suppress the useless one, we drive the proposed network to learn more robust feature representation. Finally, the experiments demonstrate that our method achieves competitive performance on widely used datasets of video salient object detection and proves the effectiveness of the proposed components.

\bibliographystyle{named}
\bibliography{ijcai21}

\end{document}